\documentclass[letterpaper, 10 pt, conference]{ieeeconf}  %

\IEEEoverridecommandlockouts                              %

\usepackage{graphicx} %
\usepackage[utf8]{inputenc}

\usepackage[
backend=biber,
style=ieee,
sorting=ynt,
doi=false,isbn=false,url=false
]{biblatex}
 
\title{\LARGE \bf
Active Perception and Representation for Robotic Manipulation
}

\author{Youssef Zaky $^{1, 2}$ , Gaurav Paruthi $^{1}$, Bryan Tripp $^{2}$, James Bergstra $^{1}$%
\thanks{$^{1}$ Kindred Systems}%
\thanks{$^{2}$ Centre for Theoretical Neuroscience, University of Waterloo}%
}

\begin{document}

\maketitle
\thispagestyle{empty}
\pagestyle{empty}

\begin{abstract}
The vast majority of visual animals actively control their eyes, heads, and/or bodies to direct their gaze toward different parts of their environment \cite{land1999}. In contrast, recent applications of reinforcement learning in robotic manipulation employ cameras as passive sensors. These are carefully placed to view a scene from a fixed pose. Active perception allows animals to gather the most relevant information about the world and focus their computational resources where needed. It also enables them to view objects from different distances and viewpoints, providing a rich visual experience from which to learn abstract representations of the environment. Inspired by the primate visual-motor system, we present a framework that leverages the benefits of active perception to accomplish manipulation tasks. Our agent uses viewpoint changes to localize objects, to learn state representations in a self-supervised manner, and to perform goal-directed actions. We apply our model to a simulated grasping task with a 6-DoF action space. Compared to its passive, fixed-camera counterpart, the active model achieves 8\% better performance in targeted grasping. Compared to vanilla deep Q-learning algorithms \cite{Quillen}, our model is at least four times more sample-efficient, highlighting the benefits of both active perception and representation learning.
\end{abstract}

\section{INTRODUCTION}

Vision-based deep reinforcement learning has recently been applied to robotic manipulation tasks with promising success (\cite{Quillen, Kalashnikov, haarnoja2, ebert2018, Agrawal2016, Schwab}). Despite successes in terms of task performance, reinforcement learning is not an established solution method in robotics, mainly because of lengthy training times (e.g., four months with seven robotic arms in \cite{Kalashnikov}). We argue in this work that reinforcement learning can be made much faster, and therefore more practical in the context of robotics, if additional elements of human physiology and cognition are incorporated: namely the abilities associated with active, goal-directed perception.

\begin{figure}[t!]
  \centering
  \includegraphics[width=0.9\columnwidth]{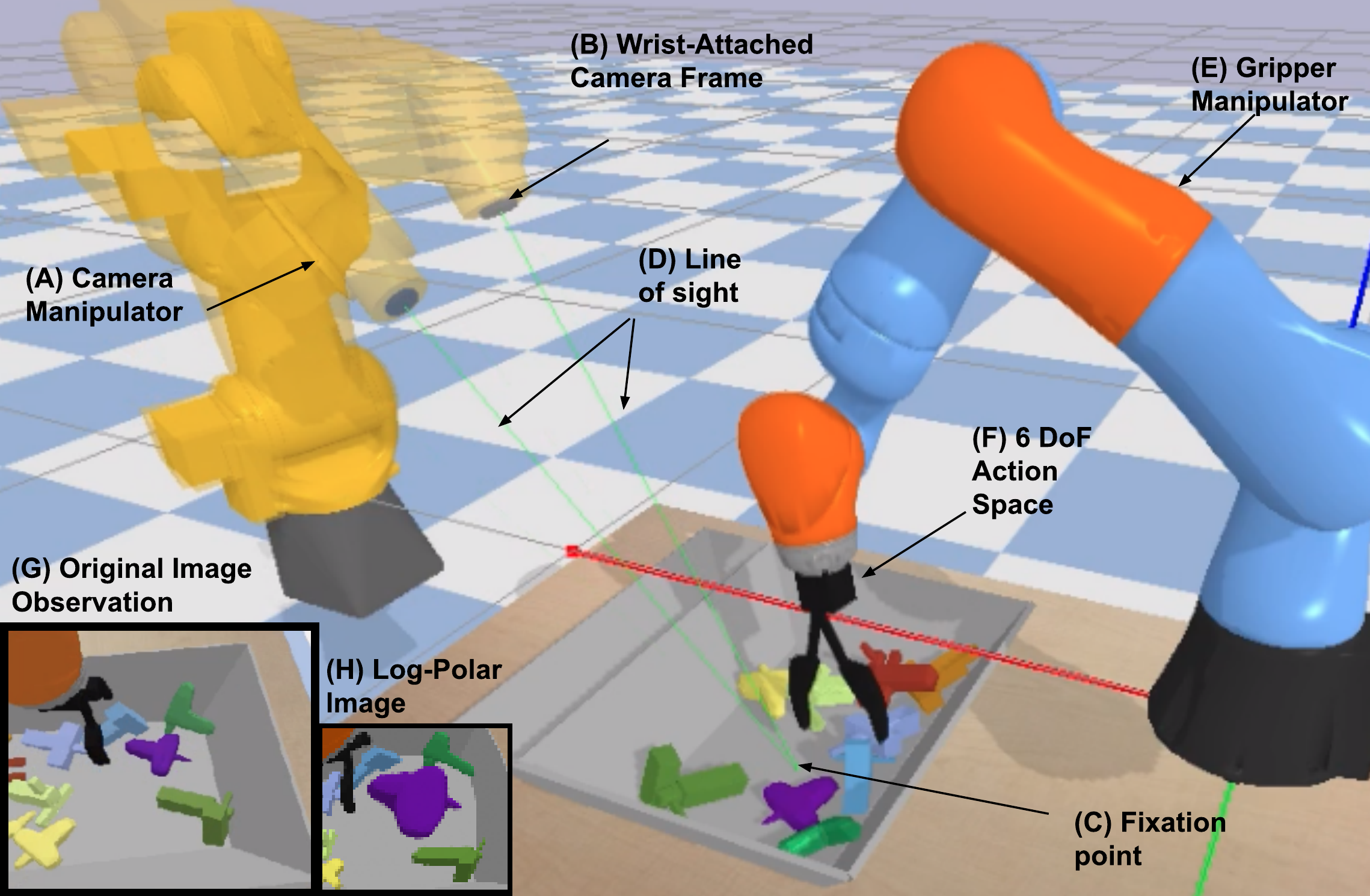}
  \caption{Our active perception setup, showing the interaction between two manipulators (A, E). The camera manipulator (A) is used to shift a wrist-attached camera frame (B) about a fixation point (C) while maintaining the line of sight (D) aligned with the point. The gripper manipulator (E) is equipped with a 6-DoF action space. The original image observation (G) is sampled with a log-polar like transform to obtain (H). Note that the log-polar sampling reduces the image size by a factor of four (256x256 to 64x64) without sacrificing the quality of the central region.}
\end{figure}

We focus in particular on two related strategies used by the human visual system \cite{Gegenfurtner2016}.
First, the human retina provides a space-variant sampling of the visual field such that the density of photoreceptors is highest in the central region (fovea) and declines towards the periphery. This arrangement allows humans to have high-resolution, sharp vision in a small central region while maintaining a wider field-of-view. Second, humans (and primates in general) possess a sophisticated repertoire of eye and head movements \cite{Liversedge2011} that align the fovea with different visual targets in the environment (a process termed `foveation`). This ability is essential for skillful manipulation of objects in the world: under natural conditions, humans will foveate an object before manipulating it \cite{Johansson2001}, and performance declines for actions directed at objects in the periphery \cite{Prado2005}. 

These properties of the primate visual system have not gone unnoticed in the developmental robotics literature. Humanoid prototypes are often endowed with viewpoint control mechanisms (\cite{Orabona, Colombo1996, Metta2000, Falotico2009}). The retina-like, space-variant visual sampling is often approximated using the log-polar transform, which has been applied to a diverse range of visual tasks (see \cite{Traver2010} for a review). Space-variant sampling, in conjunction with an active perception system, allows a robot to perceive high-resolution information about an object (e.g., shape and orientation) and still maintain enough contextual information (e.g., location of object and its surroundings) to produce appropriate goal-directed actions. We mimic these two properties in our model. First, in addition to the grasping policy, we learn an additional `fixation' policy that controls a second manipulator (Figure 1A, B) to look at different objects in space. Second, images observed by our model are sampled using a log-polar like transform (Figure 1G, H), disproportionately representing the central region.

Active perception provides two benefits in our model: an attention mechanism (often termed `hard` attention in deep learning literature) and an implicit way to define goals for downstream policies (manipulate the big central object in view). A third way we exploit viewpoint changes is for multiple-view self-supervised representation learning. The ability to observe different views of an object or a scene has been used in prior work (\cite{Sermanet2017a, Eslami2018, Dwibedi2018, Yan2017a}) to learn low-dimensional state representations without human annotation. Efficient encoding of object and scene properties from high-dimensional images is essential for vision-based manipulation; we utilize Generative Query Networks \cite{Eslami2018} for this purpose. While prior work assumed multiple views were available to the system through unspecified or external mechanisms, here we use a second manipulator to change viewpoints and to parameterize camera pose with its proprioceptive input.

We apply our active perception and representation (APR) model to the benchmark simulated grasping task published in \cite{Quillen}. We show that our agent can a) identify and focus on task-relevant objects, b) represent objects and scenes from raw visual data, and c) learn a 6-DoF grasping policy from sparse rewards. In both the 4-DoF and 6-DoF settings, APR achieves competitive performance (85\% success rate) on test objects in under 70,000 grasp attempts, providing a significant increase in sample-efficiency over algorithms that do not use active perception or representation learning \cite{Quillen}. Our key contributions are:
\begin{itemize}
\item a biologically inspired model for visual perception applied to robotic manipulation
\item a simple approach for joint learning of eye and hand control policies from sparse rewards
\item a method for sample-efficient learning of 6-DoF, viewpoint-invariant grasping policies
\end{itemize}

\begin{figure*} [t!]
  \centering
  \includegraphics[width=0.9\textwidth]{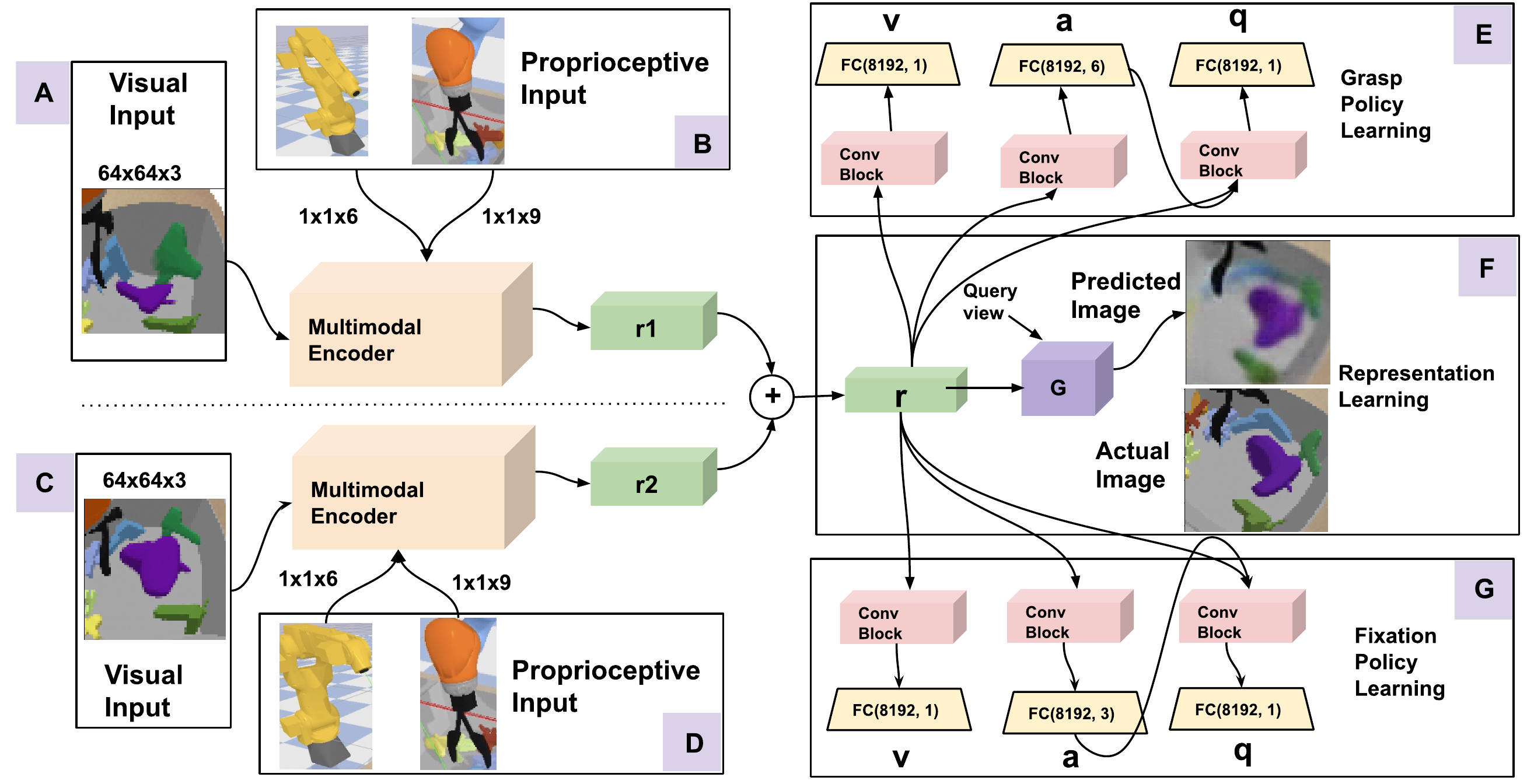}
  \caption{The APR Model. Visual (A) and proprioceptive (B) input from one view are encoded by the multimodal encoder to obtain the representation $r1$. The representation $r2$ is similarly obtained by encoding the visual (C) and proprioceptive input (D) from a second view. $r1$ and $r2$ are added to obtain the combined scene representation $r$. The action $a$, state-value $v$, and action-value function $q$ are computed for both the grasp policy (E) and the fixation policy (G). The GQN generator $g$ predicts the image from a query viewpoint, which is compared to the ground truth image from that view (F). Yellow boxes represent fully connected layers. Pink boxes represent convolutional blocks.}
  \label{figure:model}
\end{figure*}
\section{Related Work}

\subsection{Deep RL for Vision-Based Robotic Manipulation} 

Our task is adapted from the simulated setup used in \cite{Quillen} and \cite{Kalashnikov}. \cite{Kalashnikov} showed that intelligent manipulation behaviors can emerge through large-scale Q-learning in simulation and on real world robots. The robots were only given RGB inputs from an uncalibrated camera along with proprioceptive inputs. A comparative study of several Q-learning algorithms in simulation was performed in \cite{Quillen} using the same task setup. Achieving success rate over 80\% required over 100K grasp attempts. Performance of 85\% or over is reported with 1M grasp attempts. Furthermore, \cite{Kalashnikov} and \cite{Quillen} restricted the action space to 4-DoF (top-down gripper orientations). We remove such restrictions, allowing the gripper to control all 6-DoF as this is important for general object manipulation. 

Reinforcement learning with high-dimensional inputs and sparse rewards is data intensive (\cite{Mnihb, Kaiser2019}), posing a problem for real world robots where collecting large amounts of data is costly. Goal-conditioned policies have been used to mitigate the sparse reward problem in previous work (\cite{Andrychowicz, Nair2018}). In addition to optimizing the sparse rewards available from the environments, policies are also optimized to reach different goals (or states), providing a dense learning signal. We adopt a similar approach by using the 3D points produced by a fixation policy as reaching targets for the grasping policy. This ensures that the grasping policy always has a dense reward signal. We use the Soft Actor-Critic algorithm \cite{Haarnoja2018} for policy learning, which was shown to improve both sample-efficiency and performance on real world vision-based robotic tasks \cite{haarnoja2}. 

\subsection{Multiple View Object and Scene Representation Learning}

Classical computer vision algorithms infer geometric structure from multiple RGB or RGBD images. For example, structure from motion \cite{Ozyesil2017} algorithms use multiple views of a scene across time to produce an \textit{explicit} representation of it in the form of voxels or point sets. Multiple, RGBD images across space can also be integrated to produce such explicit representations \cite{Zollhofer2018}. The latter approach is often used to obtain a 3D scene representation in grasping tasks (\cite{Zeng2017a, Zeng2017}). In contrast to these methods, neural-based algorithms learn \textit{implicit} representations of a scene. This is typically structured as a self-supervised learning task, where the neural network is given observations from some viewpoints and is tasked with predicting observations from unseen viewpoints. The predictions can take the form of RGB images, foreground masks, depth maps, or voxels (\cite{Rezende2016, Gadelha, Tulsiani2017, Yan2016, Wu2016b, Eslami2018}). The essential idea is to infer low-dimensional representations by exploiting the relations between the 3D world and its projections onto 2D images. A related approach is described in \cite{Florence} where the network learns object descriptors using a pixelwise contrastive loss. However, data collection required a complex pre-processing procedure (including a 3D reconstruction) in order to train the network in the first place. Instead of predicting observations from a different views, Time Contrastive Networks (TCNs) \cite {Sermanet2017a} use a metric learning loss to embed different viewpoints closer to each other than to their temporal neighbors, learning a low-dimensional image embedding in the process.

Multiple view representation learning has proven useful for robotic manipulation. TCNs \cite{Sermanet2017a} enabled reinforcement learning of manipulation tasks and imitation learning from humans. Perspective transformer networks \cite{Yan2016} were applied to a 6-DoF grasping task in \cite{Yan2017a}, showing improvements over a baseline network. \cite{Florence} used object descriptors to manipulate similar objects in specific ways. GQNs \cite{Eslami2018} were shown to improve data-efficiency for RL on a simple reaching task. In this work we chose to use GQNs for several reasons: a) they require minimal assumptions, namely, the availability of RGB images only and b) they can handle unstructured scenes, representing both multiple objects and background, contextual information. We adapted GQNs to our framework in three ways. First, viewpoints are not arbitrarily distributed across the scene, rather they maintain the line of sight directed at the 3D point chosen by the fixation policy. Second, we apply the log-polar like transform to all the images, such that the central region of the image is disproportionately represented. These two properties allow the representation to be largely focused on the central object, with contextual information receiving less attention according to its distance from the image center. Third, instead of learning the representation prior to the RL task as done in \cite{Eslami}, we structure the representation learning as an auxiliary task that it is jointly trained along with the RL policies. This approach has been previously used in \cite{Jaderberg2016a} for example, resulting in 10x better data-efficiency on Atari games. Thus APR jointly optimizes two RL losses and a representation loss from a single stream of experience. 

\subsection{Visual Attention Architectures}

Attention mechanisms are found in two forms in deep learning literature \cite{Xu2015a}. ``Soft'' attention is usually applied as a weighting on the input, such that more relevant parts receive heavier weighting. ``Hard'' attention can be viewed as a specific form of soft attention, where only a subset of the attention weights are non-zero. When applied to images, this usually takes the form of an image crop. Hard attention architectures are not the norm, but they have been used in several prior works, where a recurrent network is often used to  iteratively attend to (or ``glimpse'') different parts of an image. In \cite{Eslami}, this architecture was used for scene decomposition and understanding using variational inference. In \cite{Gregor2015}, it was used to generate parts of an image one at a time.  In \cite{Mnih}, it was applied to image classification tasks and dynamic visual control for object tracking. More recently in \cite{Elsayed}, hard attention models have been significantly improved to perform image classification on ImageNet. Our work can be seen as an extension of these architectures from 2D to 3D. Instead of a 2D crop, we have a 3D shift in position and orientation of the camera that changes the viewpoint. We found a single glimpse was sufficient to reorient the camera so we did not use a recurrent network for our fixation policy. 

\section{Method}

\subsection{Overview}

We based our task on the published grasping environment \cite{Quillen}. A robotic arm with an antipodal gripper must grasp procedurally generated objects from a tray (Figure 1). We modify the environment in two ways: a) the end-effector is allowed to move in full 6-DoF (as opposed to 4-DoF), and b) a second manipulator (the head) is added with a camera frame fixed onto its wrist. This second manipulator is used to change the viewpoint of the attached camera. The agent therefore is equipped with two action spaces: a viewpoint control action space and a grasp action space. Since the camera acts as the end-effector on the head, its position and orientation in space are specified by the joint configuration of that manipulator: $v = (j_1, j_2, j_3, j_4, j_5, j_6)$. The viewpoint action space is three-dimensional, defining the point of fixation $(x, y, z)$ in 3D space. Given a point of fixation, we sample a viewpoint from a sphere centered on it. The yaw, pitch and distance of the camera relative to the fixation point are allowed to vary randomly within a fixed range. We then use inverse kinematics to move the head to the desired camera pose. Finally, the second action space is 6-dimensional $(dx, dy, dz, da, db, dc)$, indicating the desired change in gripper position and orientation (Euler angles) at the next timestep. 

Episodes are structured as follows. The agent is presented with an initial view (fixation point at the center of the bin) and then executes a glimpse by moving its head to fixate a different location in space. This forms a single-step episode from the point of view of the glimpse policy (which reduces the glimpse task to the contextual bandits formulation). The fixation location is taken as the reaching target; this defines the auxiliary reward for the grasping policy. The grasping policy is then executed for a fixed number of timesteps (maximum 15) or until a grasp is initiated (when the tool tip drops below a certain height). This defines an episode from the point of view of the grasping policy. The agent receives a final sparse reward if an object is lifted and the tool position at grasp initiation was within 10cm of the fixation target. The latter condition encourages the agent to look more precisely at objects, as it is only rewarded for grasping objects it was looking at. The objective of the task is to maximize the sparse grasp success reward. The grasping policy is optimized using the sparse grasp reward and the auxiliary reach reward, and the fixation policy is optimized using the grasp reward only. 

Note that all views sampled during the grasping episode are aligned with the fixation point. In this manner, the grasping episode is implicitly conditioned by the line of sight. Essentially, this encourages the robot to achieve a form of eye-hand coordination where reaching a point in space is learnt as a reusable skill. The manipulation task is thus decomposed into two problems: localize and fixate a relevant object, then reach for and manipulate said object. 

\subsection{Model}

An overview of APR is given in Figure \ref{figure:model}. Multimodal input from one view, consisting of the view parameterization (six joint angles of the head $v = (j_1, j_2, j_3, j_4, j_5, j_6)$), image ($64 \times 64 \times 3$) and gripper pose $g = (x, y, z, \sin(a), \cos(a), \sin(b), \cos(b), \sin(c), \cos(c))$, is encoded into a scene representation, $r1$, using a seven layer convolutional network with skip connections. $(a, b, c)$ are the Euler angles defining the orientation of the gripper. The scene representation $r1$ is of size $16 \times 16 \times 256$. The proprioceptive input vectors $g$ and $v$ are given spatial extent and tiled across the spatial dimension ($16 \times 16$) before being concatenated to an intermediate layer of the encoder. The input from a second view (Figure 2C, D) is similarly used to obtain $r2$, which is then summed to obtain $r$, the combined scene representation.

The fixation policy and grasping policies operate on top of $r$. Their related outputs (action $a$, state-value $v$ and action-value functions $q$) are each given by a convolutional block followed by a fully-connected layer. The convolutional blocks each consist of three layers of $3 \times 3$ kernels with number of channels 128, 64, and 32 respectively. The generator is a conditional, autoregressive latent variable model that uses a convolutional LSTM layer. Conditioned on the representation $r$, it performs 12 generation steps to produce a probability distribution over the query image. The encoder and generator architecture are unmodified from the original work, for complete network details we refer the reader to \cite{Eslami2018}.

\begin{figure}[t!] 
  \centering
  \includegraphics[width=0.9\columnwidth]{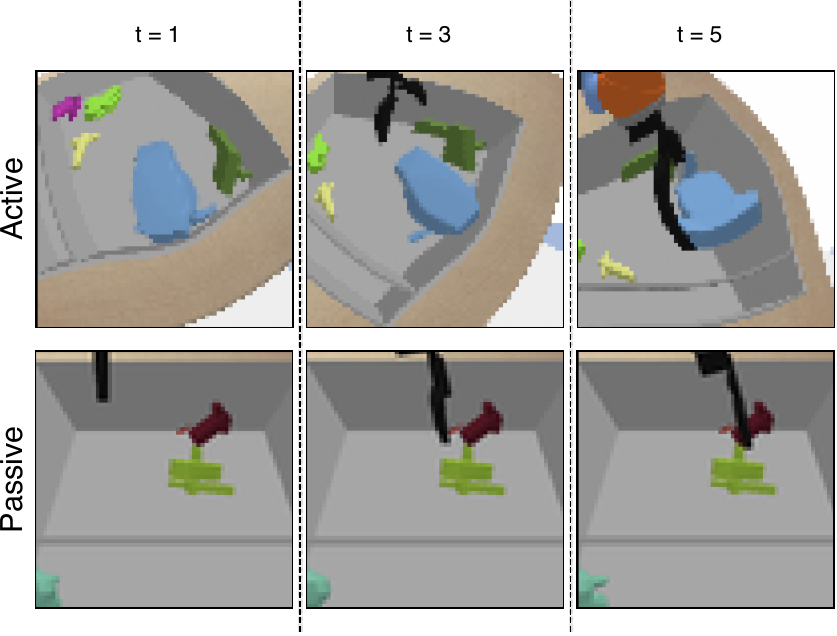}
  \caption{Comparing visual inputs of the active and passive models during a five step episode. Top: images sampled from different views centered on the target object. Bottom: images from one of the static cameras of the passive model. An interesting feature of the active input is that the gripper appears larger as it approaches the target object, providing an additional learning cue.}
\end{figure}

The log-polar like sampling we use is defined as follows. Let $(u, v) \in [-1, 1] \times [-1, 1]$ be a coordinate in a regularly spaced image sampling grid. We warp $(u, v)$ to obtain the log-polar sampling coordinate $(u', v')$ using the following equation: 

$$
(u', v') = \log(\sqrt{u^2 + v^2} + 1) \cdot (u, v)
$$

\subsection{Learning}

We learn both policies using the Soft Actor-Critic algorithm \cite{Haarnoja2018}, which optimizes the maximum-entropy RL objective. For detailed derivations of the loss functions for policy and value learning we refer the reader to \cite{Haarnoja2018}. In conjunction with the policy learning, the multimodal encoder and generator are trained using the generative loss (evidence lower bound) of GQN. This loss consists of KL divergence terms and a reconstruction error term obtained from the variational approximation \cite{Eslami2018}. Note that the encoder does not accumulate gradients from the reinforcement learning losses and is only trained with the generative loss. To obtain multiple views for training, we sample three viewpoints centered on the given fixation point at every timestep during a grasping episode. Two of those are randomly sampled and used as context views to obtain $r$, the third acts as the ground truth for prediction. We did not perform any hyperparameter tuning for the RL or GQN losses and used the same settings found in \cite{Haarnoja2018} and \cite{Eslami2018}. 

\section{Experiments}

We perform three experiments that examine the performance of active vs passive models (Section A), of active models that choose their own targets (Section B), and the benefits of log-polar images and representation learning for active models (Section C).

In our experiments, training occurs with a maximum of 5 objects in the bin. A typical run takes approximately 26 hours (on a single machine), with learning saturating before 70K grasps. Every episode, the objects are sampled from a set of 900 possible training objects. For evaluation, we use episodes with exactly 5 objects present in the bin. Evaluation objects are sampled from a different set of 100 objects, following the protocol in \cite{Quillen}. 

\subsection{Active vs Passive Perception}

We evaluate active looking vs a passive (fixed) gaze at targeted grasping. This setting is designed to test goal-oriented manipulation, rather than grasping any object arbitrarily. For this experiment, we do not learn a fixation policy, but instead use goals, or target objects, that are selected by the environment. The policies are only rewarded for picking up the randomly chosen target object in the bin. The active model and the passive model receive the same inputs (visual and proprioceptive) along with a foreground object mask that indicates the target object. The only difference between the two models is the nature of the visual input: the active model observes log-polar images that are centered on the target object, while the passive model observes images of the bin from three static cameras (Figure 3). The static cameras are placed such that each can clearly view the contents of the bin and the gripper. This mimics a typical setup where cameras are positioned to view the robot workspace, with no special attention to any particular object or location in space. Using an instance mask to define the target object was previously done in \cite{Fangb} for example. Note that the generator is trained to also reconstruct the mask in this case, forcing the representation $r$ to preserve the target information. 

Table 1 (Active-Target, Passive-Target) shows the evaluation performance of the active vs passive model with environment selected targets. We observe that the active model achieves 8\% better performance. Figure 4 (yellow vs blue curves) shows that the active model is more sample-efficient as well. 

The performance of the passive model (at 76\%) is in line with the experiments in \cite{Quillen} on targeted grasping. All algorithms tested did not surpass an 80\% success rate, even with 1M grasp attempts. The experiment above suggests that, had the robot been able to observe the environment in a more ``human-like'' manner, targeted grasping performance could approach performance on arbitrary object grasping.

\begin{table}[t]
\caption{Evaluation Performance}
\label{table_example}
\begin{center}
\begin{tabular}{|c||c|}
\hline
Model & Grasp Success Rate\\
\hline
Active-Target & 84\% \\
\hline
Passive-Target & 76\% \\
\hline
Active-Target w/o Log-Polar & 79\% \\
\hline
Active-Learned-6D (after 70K grasps)& 85\% \\
\hline
Active-Learned-4D (after 25K grasps) & 85\% \\
\hline
\end{tabular}
\end{center}
\end{table}

\begin{figure}[t!] 
  \centering
  \includegraphics[width=0.9\columnwidth]{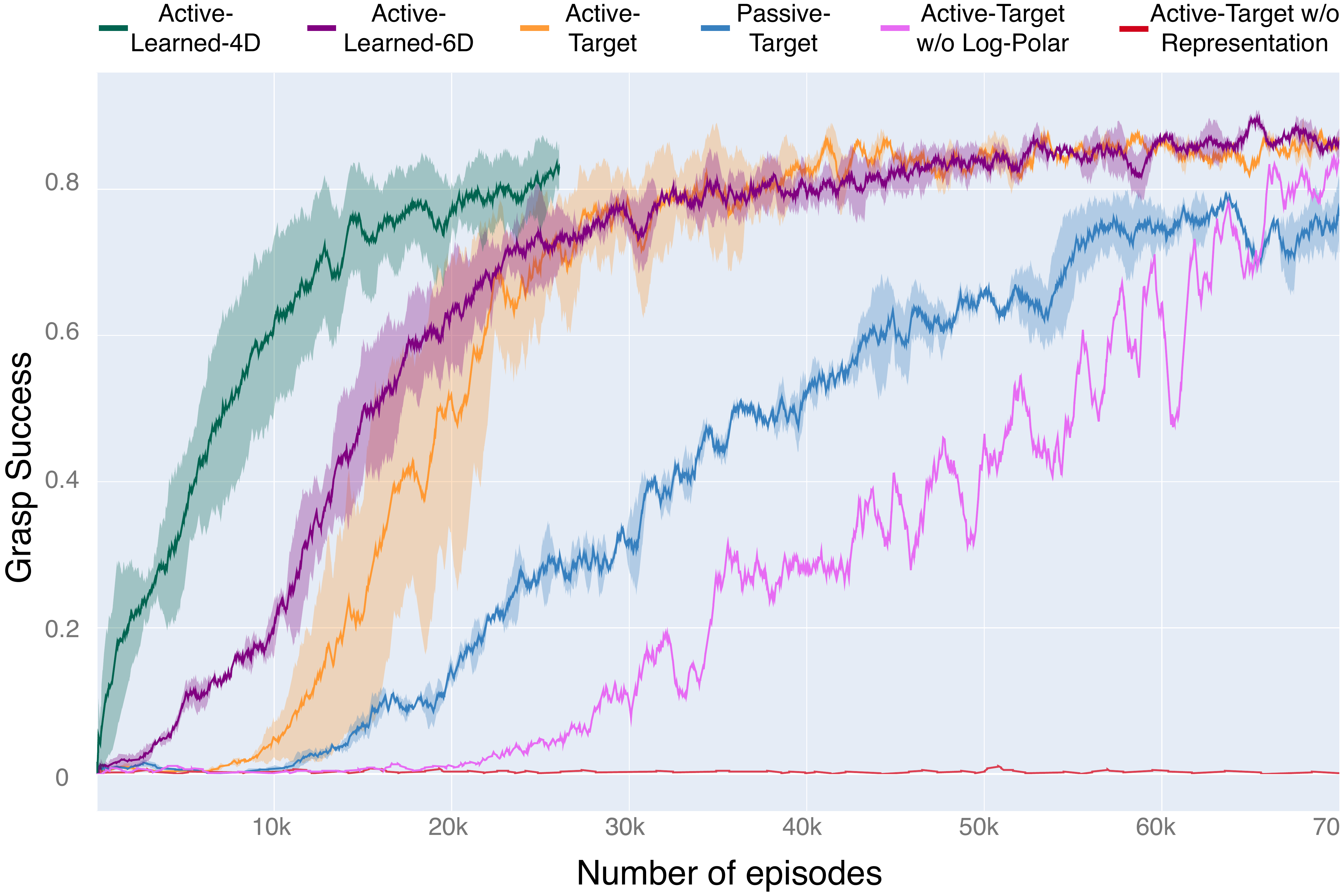}
  \caption{Learning curves for our experiments. Active-Target, Passive-Target: active and passive models with environment selected targets. Active-Learned: full APR model with fixation policy. Active-w/o representation, Active-Target w/o log-polar: APR versions without representation learning or log-polar sampling, respectively. Shaded regions indicate the standard deviation over two independent runs with different random seeds.}
\end{figure} 

\subsection{Learning Where to Look}

The experiment above shows that active perception outperforms passive models in goal-oriented grasping. But can a robot learn where to look? Here we use the full version of the model with the learned fixation policy. Grasp rewards are given for picking up any object, as long as the object was within 10cm of the fixation point. This ensures that the model is only rewarded for goal-oriented behavior. In this setting, the model learns faster in the initial stages than in the targeted grasping case (Figure 4) and is slightly better in final performance (Table 1). This does not necessarily imply that this model is better at grasping, it could be due to the model choosing easier grasping targets. The latter may nevertheless be a virtue depending on the context (e.g., a bin-emptying application). This result indicates that active perception policies can be learnt in conjunction with manipulation policies. 

Note that the full version of APR does not use additional information from the simulator beyond the visual and proprioceptive inputs. In contrast to Section A, the fixation point (and therefore the auxiliary reaching reward), is entirely self-generated. This makes APR directly comparable to the vanilla deep Q-learning learning algorithms studied in \cite{Quillen}. With  100K  grasp  attempts,  the  algorithms  in  \cite{Quillen}  achieve approximately  80\% success  rate.  We tested the model in the 4-DoF case, where it achieves an 85\% success rate with 25K grasps (Table 1).  Therefore, APR outperforms these previous baselines with four times fewer samples. [33] reported improved success rates of 89-91\% with vanilla deep Q-learning after 1M  grasps (though it was not reported what levels of performance were attained between 100K and 1M grasps). On the more challenging 6-DoF version, we  achieve  an  85\%  success  rate  with  70K grasps, but we have not yet extended the simulations to 1M grasps to allow a direct comparison with these results. 

\subsection{Ablations} 

To examine effects of the log-polar image sampling and the representation learning, we ran two ablation experiments in the environment selected target setting (as in Section A). Figure 4 (red curve) shows that APR without representation learning achieves negligible improvement within the given amount of environment interaction. (Without the representation learning loss, we allow the the RL loss gradients to backpropagate to the encoder, otherwise it would not receive any gradient at all). The pink curve shows APR without log-polar images. The absence of the space-variant sampling impacts both the speed of learning and final performance (Table 1).

\section{Discussion and Future Work}

We presented an active perception model that learns where to look and how to act using a single reward function. We showed that looking directly at the target of manipulation enhances performance compared to statically viewing a scene (Section 4A), and that our model is competitive with prior work while being significantly more data-efficient (Section 4B). We applied the model to a 6-DoF grasping task in simulation, which requires appropriate reaching and object maneuvering behaviors. This is a more challenging scenario as the state space is much larger than the 4-DoF state space that has typically been used in prior work (\cite{Pinto2015, Quillen, Kalashnikov}). 6-DoF control is necessary for more general object manipulation beyond top-down grasping. Figure 5 shows interesting cases where the policy adaptively orients the gripper according to scene and object geometry. 

The biggest improvement over vanilla model-free RL algorithms came from representation learning, which benefited both passive and active models. Figure 6 shows sample generations from query views along with ground truth images from a single run of the active model. Increasingly sharp renderings (a reflection of increasingly accurate scene representation) correlated with improving performance as the training run progressed. While the generated images retained a degree of blurriness, the central object received a larger degree of representational capacity simply by virtue of its disproportionate size in the image. This is analogous to the phenomenon of ``cortical magnification'' observed in visual cortex, where stimuli in the central region of the visual field are processed by a larger number of neurons compared to stimuli in the periphery \cite{DANIEL1961}. We suspect that such a representation learning approach -- one that appropriately captures the context, the end-effector, and the target of manipulation -- is useful for a broad range of robotic tasks.  

\begin{figure}[t] 
  \centering
  \includegraphics[width=0.9\columnwidth]{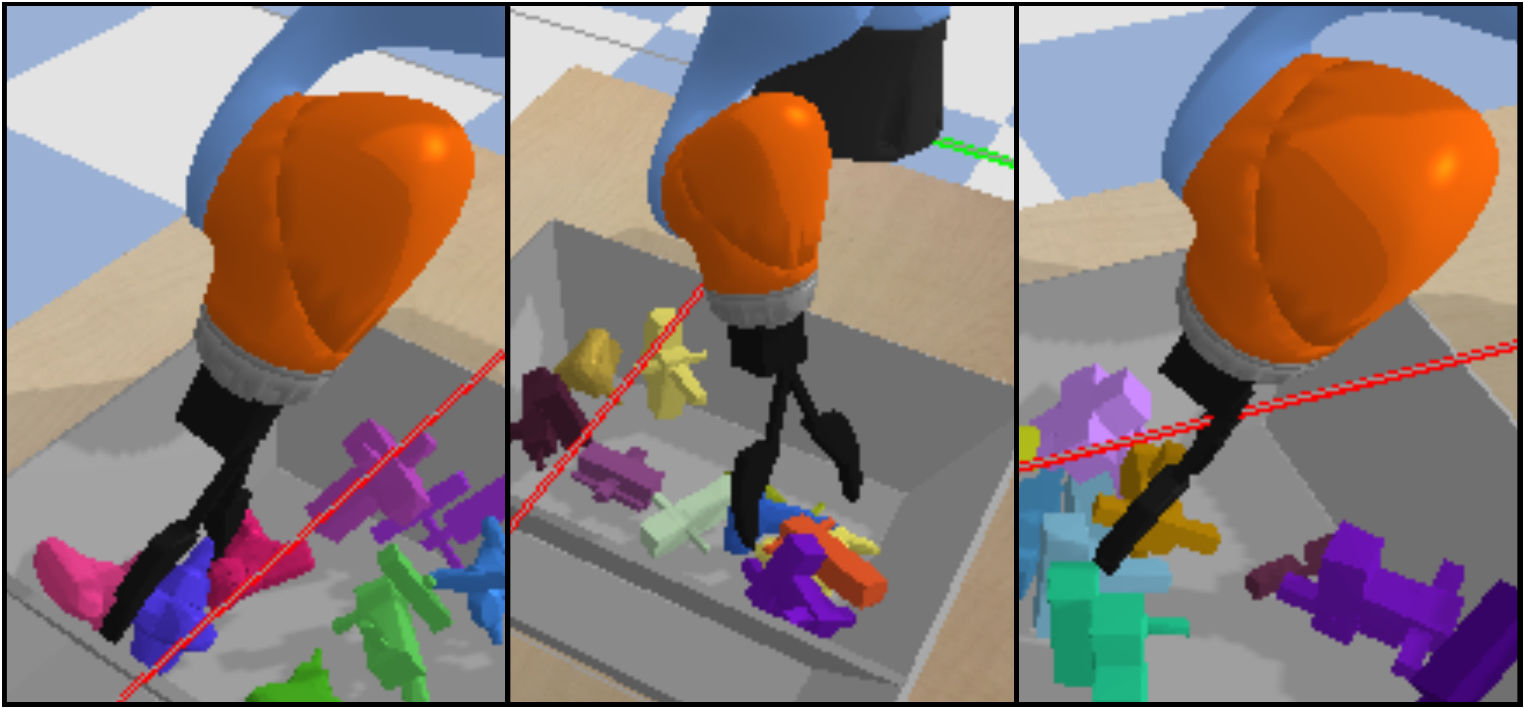}
  \caption{Examples of pre-grasp orienting behaviors due to the policy's 6-DoF action space.}
\end{figure}

\begin{figure}[t!]
  \centering
  \includegraphics[width=0.9\columnwidth]{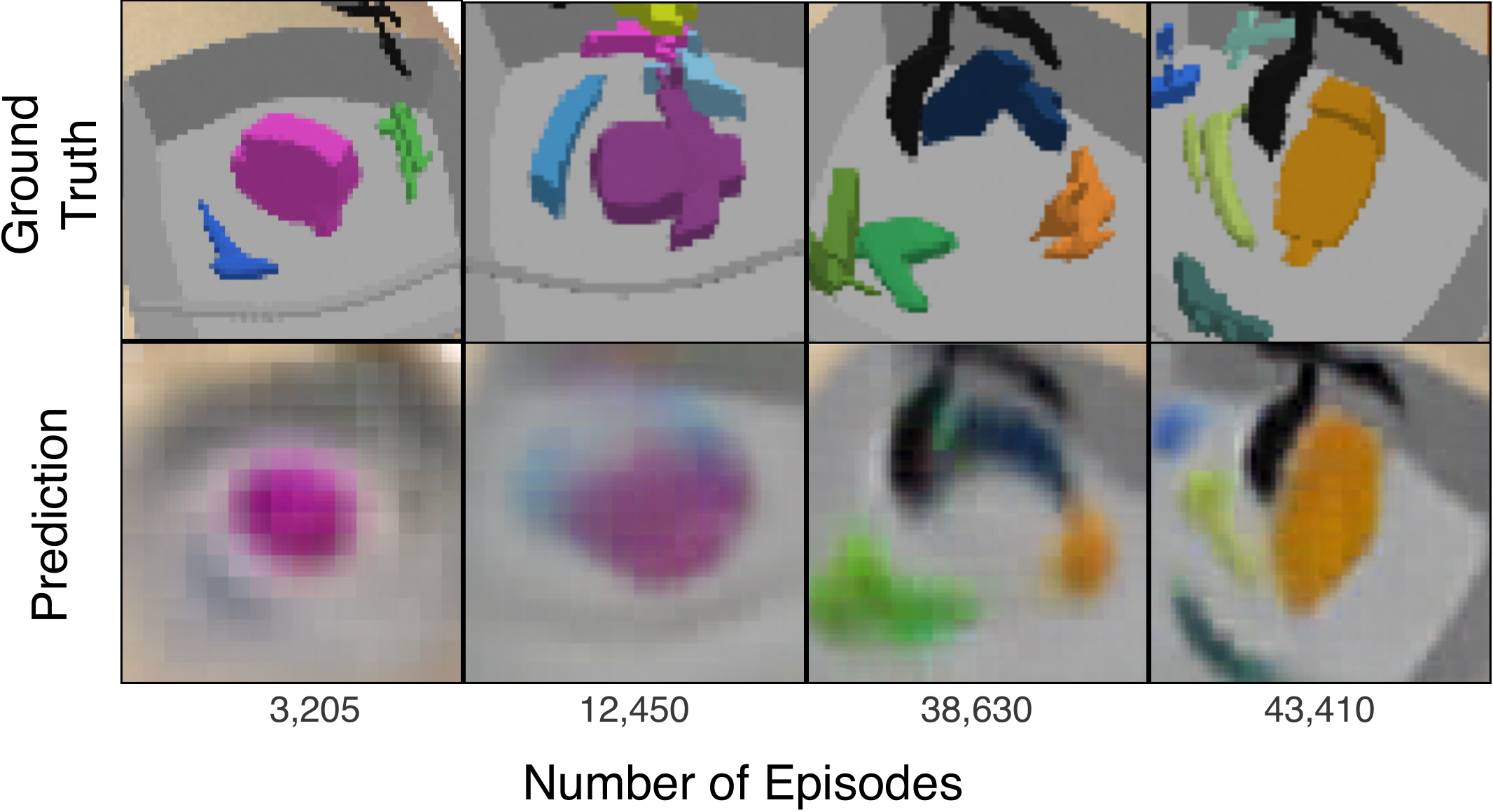}
  \caption{Scene renderings from query views at different snapshots during active model training. At later stages, the gripper, central object, and bin are well-represented. Surrounding objects occupy fewer pixels in the image, so they are not represented in as much detail.}
\end{figure} 

Looking ahead to testing the APR model in a physical environment, we see additional challenges. Realistic images may be more difficult for the generative model of GQN, which could hamper the representation learning. Exploration in a 6-DoF action space is more time-consuming and potentially more collision-prone than a top-down, 4-DoF action space. Some mechanism for force sensing or collision avoidance might be needed to prevent the gripper from colliding with objects or the bin. Active camera control introduces another complicating factor. It requires a physical mechanism to change viewpoints and a way of controlling it. We used a second 6-DoF manipulator in our simulator, but other simpler motion mechanisms are possible. Learning where to look with RL as we did in this work may not be necessary. It might be possible to orient the camera based on 3D location estimates of relevant targets.
 
Looking at relevant targets in space and reaching for them are general skills that serve multiple tasks. We believe an APR-like model can therefore be be applied to a wide range of manipulation behaviors, mimicking how humans operate in the world. Whereas we structured how the fixation and grasping policies interact (``look before you grasp''), an interesting extension is where both policies can operate dynamically during an episode. For example, humans use gaze-shifts to mark key positions during extended manipulation sequences \cite{Johansson2001}. In the same manner that our fixation policy implicitly defines a goal, humans use sequences of gaze shifts to indicate subgoals and monitor task completion \cite{Johansson2001}. The emergence of sophisticated eye-hand coordination for object manipulation would be exciting to see.

\section{Conclusion}

\cite{Hassabis2017} argues that neuroscience (and biology in general) still contain important clues for tackling AI problems. We believe the case is even stronger for AI in robotics, where both the sensory and nervous systems of animals can provide a useful guide towards intelligent robotic agents. We mimicked two central features of the human visual system in our APR model: the space-variant sampling property of the retina, and the ability to actively perceive the world from different views. We showed that these two properties can complement and improve state-of-the-art reinforcement learning algorithms and generative models to learn representations of the world and accomplish challenging manipulation tasks efficiently. Our work is a step towards robotic agents that bridge the gap between perception and action using reinforcement learning.

\printbibliography
\end{document}